\pdfoutput=1

\documentclass[11pt]{article}

\usepackage[final]{acl}
\usepackage{times}
\usepackage{latexsym}

\usepackage[T1]{fontenc}
\usepackage[utf8]{inputenc}

\usepackage{amsmath}

\usepackage{microtype}
\usepackage{inconsolata}
\usepackage{enumitem}

\usepackage{graphicx}
\usepackage{caption}
\usepackage{booktabs}

\usepackage{color}
\usepackage{xcolor}
\definecolor{light-gray}{gray}{0.95}
\usepackage{listings}
\lstdefinestyle{promptstyle}{
  backgroundcolor=\color{light-gray}, 
  basicstyle=\small\ttfamily,       
  breakatwhitespace=false,         
  breaklines=true,                 
  captionpos=b,                    
  keepspaces=true,                 
  showspaces=false,                
  showstringspaces=false,
  showtabs=false,                  
  tabsize=2,
  frame=single,                    
  framerule=0.5pt,
  framesep=5pt,
  rulesepcolor=\color{gray}        
}

\title{Evaluating The Impact of Stimulus Quality in Investigations of LLM Language Performance}

\author{Timothy Pistotti \\
  University of Auckland \\
   \\\And
  Jason Brown \\
  University of Auckland \\
  \\\And
  Michael Witbrock\\
  University of Auckland
  }

\begin{document}
\maketitle
\begin{abstract}
Recent studies employing Large Language Models (LLMs) to test the Argument from the Poverty of the Stimulus (APS) have yielded contrasting results across syntactic phenomena. This paper investigates the hypothesis that characteristics of the stimuli used in recent studies, including lexical ambiguities and structural complexities, may confound model performance. A methodology is proposed for re-evaluating LLM competence on syntactic prediction, focusing on GPT-2. This involves: 1) establishing a baseline on previously used (both filtered and unfiltered) stimuli, and 2) generating a new, refined dataset using a state-of-the-art (SOTA) generative LLM (Gemini 2.5 Pro Preview) guided by linguistically-informed templates designed to mitigate identified confounds. Our preliminary findings indicate that GPT-2 demonstrates notably improved performance on these refined PG stimuli compared to baselines, suggesting that stimulus quality significantly influences outcomes in surprisal-based evaluations of LLM syntactic competency. 
\end{abstract}

\section{Introduction}
\label{sec:introduction}

The Argument from the Poverty of the Stimulus (APS) remains a central topic in linguistics and cognitive science, and proposes that human linguistic competence extends beyond that supported by direct evidence available during acquisition, thereby implying contributions of innate knowledge to language learning \cite{chomsky1980rules}. Using artificial neural networks as proxies for unbiased learners, recent studies have explored the generalizations that Large Language Models (LLMs) form about linguistic phenomena. A promising line of research compares token probabilities in minimal pairs (e.g., \cite{linzen-etal-2016-assessing, futrell-etal-2019-neural, wilcox2024using, lan2024large}) following Elman’s \citeyear{elman1990finding} recommendation that language models be treated as human subjects in psycholinguistic studies.

\citet{wilcox2024using} provide significant findings in this area, demonstrating that LLMs can achieve high performance on various English filler-gap dependencies and island constraints, as measured by surprisal metrics applied to critical regions of minimal pairs of sentences \cite{wilcox2024using}. Their results challenge the necessity of linguistic innateness for these particular syntactic structures.

Building on this work, \citet{lan2024large} investigate more complex, lower frequency syntactic constructions, notably parasitic gaps (PGs) and across-the-board (ATB) movement, but argue that the observed failures of LLMs (including GPT-2) to adequately learn these structures support the APS.

This paper limits its scope to the evaluation of PGs in the context of Lan et al.'s \citeyear{lan2024large} analysis. We argue that while their work addresses crucial linguistic questions, a critical examination of their PG stimuli reveals characteristics that may interfere with LLM performance. These characteristics include: 1) unintended lexical ambiguities, 2) the structural complexity of the noun phrases hosting parasitic gaps, and 3) potential alternative repairs to ungrammaticality.

The central aim of this research is to investigate the extent to which such properties affect an LLM's predictive power in critical regions of PGs. We propose a methodology centred on generating controlled stimuli using a SOTA generative LLM (Gemini 2.5 Pro Preview) guided by precise, linguistically-informed templates. This approach seeks to mitigate the identified potential confounds while allowing for some flexibility in generation. We present preliminary findings, comparing model performance on our dataset to baselines derived from the original \citet{lan2024large} PG data. Our results suggest that stimulus quality has a significant impact on surprisal scores in critical regions, with implications for the broader APS debate and for researchers interested in applying surprisal-based methods to the investigation of LLM capabilities.

\section{Parasitic Gap Stimuli}
\label{sec:lan_stimuli_critique}

Using a Context-Free Grammar (CFG), \cite[Table 2, p.~16]{lan2024large} generated a total of 8,064 sentence tuples each comprised of $\pm Filler$, $\pm Gap$ variations, exemplified in Table \ref{tab:pg_stimuli}. While this approach allows for controlled generation, close examination reveals characteristics of the resulting PG stimuli that may influence model performance independently from the core syntactic properties of PG licensing.

\begin{table*}[!t]
    \centering
    \caption{Example paradigm for parasitic gaps. Underlined words indicate the  filler alternations. Boldfaced words indicate the critical region that shows whether the continuation is gapped or not. Reproduced from Table 4 \citet[p.~19]{lan2024large}.}
    \label{tab:pg_stimuli}
    \small
    \begin{tabular}{lp{0.44\textwidth}p{0.44\textwidth}}
        \toprule
        & \textbf{+Gap} & \textbf{-Gap} \\
        \midrule 
        \textbf{+Filler} & 
        I know \underline{who} John's talking to is about to annoy \textbf{soon}. &
        I know \underline{who} John's talking to is about to annoy \textbf{you} soon.
        \\
        \addlinespace
        
        \textbf{-Filler} & 
        I know \underline{that} John's talking to Mary is about to annoy \textbf{soon}. & 
        I know \underline{that} John's talking to Mary is about to annoy \textbf{you} soon. \\
        \bottomrule
    \end{tabular}
\end{table*}

\subsection{Unintended Ambiguity}
\label{ssec:lexical_ambiguity_lan}
This section identifies two ambiguities prevalent in Lan et al.'s PG dataset. A particularly prominent example involves the use of possessive gerunds (e.g., ``John's talking'') within the subject noun phrase (NP) that hosts the first gap (G1). 

\begin{enumerate}[label=(\arabic*), topsep=4pt, itemsep=0pt, leftmargin=2.5em]
    \item \label{ex:lan-stimulus}
    *I know who [John's talking to \_] is going to annoy you soon.
\end{enumerate}

Here, ``John's'' is ambiguous between a contraction of ``John is'' and the possessive ``John + GEN''. If interpreted as ``John is talking to \_,'' the embedded phrase might not form the intended island structure necessary for a PG, or its grammaticality profile changes. Conversely, if interpreted as a possessive, it forms the intended complex NP island. Given that the stimuli presented to the LLMs were unbracketed and unannotated (as confirmed by the project's public repository), the model must disambiguate this string without a forced reading of sentence structure. 
Similarly, constructions such as ``intent to'' (e.g., in ``I know who Bob's intent to talk to \_ is about to bother soon'' include the same ambiguity with the addition of a potential alternative rescue for the sentence's overall grammaticality (e.g., ``intent on talking to'' or ``intention of talking to'') that might alter processing ease.

\subsection{Structural Complexity of Noun Phrases}
The parasitic gap (G1) in Lan et al. stimuli is embedded within a subject NP that forms an island, derivable from their CFG rules such as ``\texttt{(NP\_COMPLEX) -> (N\_EMBEDDED) `to' (V\_EMBEDDED)}'' \cite[Table 2, p.~16]{lan2024large}, leading to structures such as the underlined portion of

\begin{enumerate}[resume, label=(\arabic*), topsep=4pt, itemsep=0pt, leftmargin=2.5em]
    \item \label{ex:complex-np}
    I know who \underline{Bob's decision to dance with \_} is likely to bother eventually.
\end{enumerate}

While subject NPs are indeed syntactic islands, the internal complexity of these specific \texttt{NP\_COMPLEX} structures (involving nominals followed by an infinitival phrase) introduces a degree of structural depth that goes beyond the more canonical adjunct PG constructions often cited as core examples in the literature \cite{culicover2001parasitic}. This complexity might itself be a confounding factor for LLMs.

\section{Method}
\label{sec:methodology}

To investigate the impact of stimulus characteristics on LLM performance for PG constructions, an experiment was designed to compare model performance across three datasets: the original Lan et al. stimuli, a filtered version of this original set, and a new, refined set generated for this study. For our analysis, we selected GPT-2 as the primary evaluation model. This choice is motivated by two factors: first, its use in both \citet{wilcox2024using} and \citet{lan2024large} provides a direct point of comparison with prior findings. Second, while GPT-2 possesses sophisticated language capabilities, it precedes the current era of massive-scale models. This makes it a more suitable testcase for hypotheses related to the Argument from the Poverty of the Stimulus, as it is less likely to have encountered rare syntactic constructions, such as parasitic gaps, at a high frequency during its training.

\subsection{Evaluation Metric: Surprisal}
Our primary measure of model performance is \textbf{surprisal}, which quantifies how unexpected a given word ($w_i$) is in its preceding context ($C$). Following standard practice \cite{wilcox2024using, lan2024large}, surprisal is calculated as the negative log probability, in bits:
\begin{equation} \label{eq:surprisal}
S(w_i \mid C) = -\log_2 P(w_i \mid C)
\end{equation}
Lower surprisal values indicate that a word is more predictable. We used this metric to calculate the $\Delta$ and Difference-in-Differences (DiD) metrics as proposed by Lan et al. (2024), where $\Delta = S(\text{-Gap Continuation}) - S(\text{+Gap Continuation})$. Model success in modelling the relevant grammaticality judgement is indicated by $\Delta_{+\text{filler}} > 0$ and $\text{DiD} = (\Delta_{+\text{filler}} - \Delta_{-\text{filler}}) > 0$.

\subsection{Datasets}
We compare GPT-2's performance across three distinct datasets for PGs:
\begin{enumerate}[label=(\alph*), topsep=4pt, itemsep=0pt, leftmargin=2em]
    \item \textbf{Original Lan et al. (2024) Stimuli:} The full dataset generated from their CFG (N=8064 items), extracted from their publicly available materials.
    \item \textbf{Filtered Lan et al. (2024) Stimuli:} A subset of the original dataset (N=5760 items) excluding all items containing the specific ambiguous constructions identified in Section~\ref{sec:lan_stimuli_critique}, namely those following the pattern: ``NAME's VERBing to''.
    \item \textbf{Refined Stimuli (This Work):} A new, controlled dataset of subject PG constructions generated using Gemini 2.5 Pro Preview (see Appendix~\ref{sec:appendix_prompt} for the full prompt template). This generation was guided by precise structural templates designed to mitigate the confounds present in the original dataset, including using unambiguous ``the [NounHead] of/about G1'' structures for the subject island, ensuring pragmatically plausible co-indexation, and using single-word critical regions for the main clause gap (G2) comparison. All such generated items underwent manual review for grammaticality.
\end{enumerate}

\subsection{Experimental Procedure}
For each dataset, we followed an identical experimental procedure:
\begin{enumerate}[label=(\arabic*), topsep=4pt, itemsep=0pt, leftmargin=2em]
    \item \textbf{Data Preprocessing:} Stimuli are formatted into a long-format CSV with columns for \texttt{sentence\_type}, \texttt{item\_id}, \texttt{condition}, and \texttt{full\_sentence}.
    \item \textbf{Surprisal Extraction:} BPE-level surprisals for each sentence are obtained from GPT-2 using a Python pipeline leveraging the \texttt{lib.py} framework from Lan et al.'s (2024) repository.
    \item \textbf{Critical Region Aggregation:} Surprisals for the single-word critical regions (the overt object NP in `-Gap` conditions or the post-gap adverb in `+Gap` conditions) are calculated by summing the surprisals of their constituent BPEs.
    \item \textbf{Analysis:} The $\Delta$ and DiD metrics, along with accuracies and one-sample t-tests, are calculated for each dataset to allow for direct comparison.
\end{enumerate}

\section{Preliminary Findings and Discussion}
\label{sec:prelim_findings}

Following the methods outlined in Section~\ref{sec:methodology}, we conducted a preliminary evaluation using GPT-2. The primary focus was to assess whether refining the stimuli for PG constructions, specifically addressing the potential confounds identified in Lan et al.'s \citeyear{lan2024large} dataset, would lead to a different pattern of performance for GPT-2.

\subsection{GPT-2 Performance on Original, Filtered, and Refined PG Stimuli}
\label{ssec:gpt2_performance_comparison}

\citet{lan2024large} originally reported that GPT-2 performed poorly on PG stimuli, with key metrics around 5.6\% accuracy for the $\Delta_{+\text{filler}} > 0$ criterion and 68.8\% accuracy for the Difference-in-Differences (DiD) criterion \citep[Figs.~5 \& 6, pp.~18,~21]{lan2024large}. This was presented as support for APS.

To establish a direct baseline, our pipeline confirmed these findings on the unfiltered original Lan et al. (2024) dataset (N=8064), yielding an accuracy of \textbf{5.61\%} for $\Delta_{+\text{filler}}>0$ and \textbf{68.75\%} for the DiD metric. Next, GPT-2's performance on the filtered subset (N=5760) was analysed. On this filtered set, accuracy for the $\Delta_{+\text{filler}}>0$ criterion improved to \textbf{7.01\%} ($\chi^2(1)=11.2381, p=0.0008$), and for the DiD criterion, accuracy improved to \textbf{72.93\%} ($\chi^2(1)=28.0780, p<0.0001$). These results provide initial empirical support for the identified constructions acting as confounds.

\begin{figure}
    \centering
    \includegraphics[width=1\linewidth]{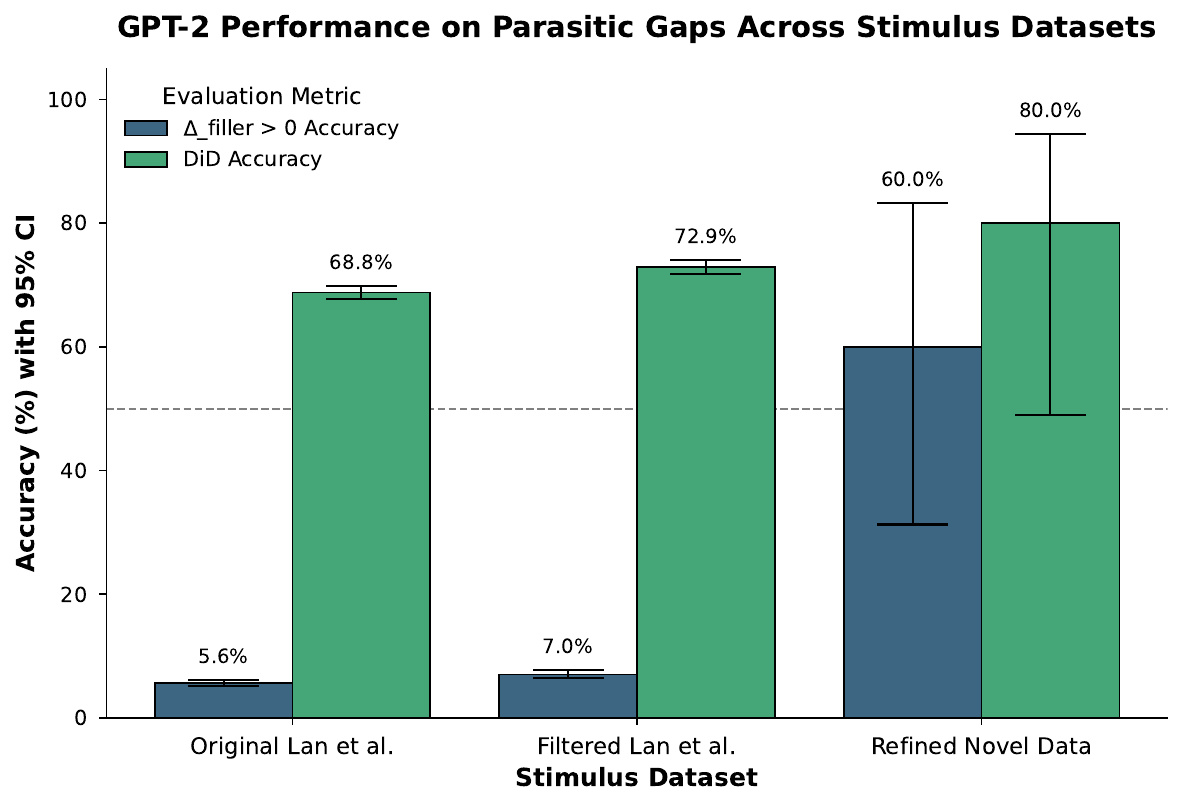}
    \caption{Comparison of GPT-2 accuracy on Parasitic Gap constructions. Accuracy is shown for the $\Delta_{+\text{filler}} > 0$ and Difference-in-Differences (DiD) > 0 criteria across the original \citep{lan2024large} dataset, a filtered version, and our own refined stimuli. Error bars represent 95\% confidence intervals.}
    \label{fig:gpt2_pg_performance}
\end{figure}

Finally, GPT-2 was evaluated on newly generated, refined \texttt{subject\_pg} stimuli (N=10 items). This yielded significant further improvement: for the $\Delta_{+\text{filler}} > 0$ metric, accuracy rose to \textbf{60.0\%} ($t(9) = 1.66, p = 0.066$, one-tailed), and for the DiD metric, accuracy reached \textbf{80.0\%}, a statistically significant effect ($t(9) = 2.64, p = 0.013$, one-tailed; 95\% CI $[0.39, 5.01]$). These comparative accuracy scores are visualized in Figure  \ref{fig:gpt2_pg_performance}.

\subsection{Discussion}

The preliminary findings from our refined \texttt{subject\_pg} dataset indicate a marked improvement in GPT-2's performance compared to the results reported by Lan et al. (2024) for their original PG stimuli using the same model. The DiD accuracy increased from $\sim$69\% to 80\%, and notably, the direct preference accuracy ($\Delta_{+\text{filler}} > 0$) jumped from $\sim$6\% to 60\%.

While these results are based on an initial set of refined stimuli and a single model, they suggest that characteristics of the test stimuli play a substantial role in LLM evaluations of complex syntax. The reduction of lexical ambiguities (like the ``John's'' issue) and the use of more canonical island structures for the G1-hosting subject NP may have allowed GPT-2 to better demonstrate any underlying sensitivity it has to PG constructions.

These findings do not nullify Lan et al.'s (2024) broader arguments regarding the APS, that more complex linguistic phenomena may be better suited to test learnability. However, they do suggest that conclusions about an LLM's failure to acquire a phenomenon might be premature if based on stimuli containing significant potential confounds. If an LLM's performance is demonstrably better on refined, unambiguous stimuli, it points to the model's sensitivity to these confounds, and implies that at least some of the previously observed ``failure'' might be attributable to the nature of the test items themselves rather than to incomplete generalization. This suggests that the introduction of unintended complexities, not directly targeted by the parasitic gap investigation, may obscure an LLM's underlying sensitivity to PG licensing, analogous to the effect of increased structural complexity (e.g., embedding depth) \citet{wilcox2024using} in reducing wh-effects in filler-gap dependencies.

The approach of using a SOTA generative LLM (Gemini 2.5 Pro Preview) guided by precise linguistic templates for creating these refined stimuli shows promise as a method for developing more robust and theoretically sound evaluation protocols. This can help in disentangling true model capabilities from noise introduced by problematic test data.

\subsection{Limitations and Future Directions}
Future work based on these preliminary findings will involve:
\begin{itemize}
    \item Expanding the refined dataset to include more items and other PG structures (e.g., adjunct PGs).
    \item Testing a wider range of LLMs, including more recent architectures and models trained on smaller datasets.
    \item Conducting a more detailed error analysis on the original Lan et al. (2024) PG dataset using our full pipeline to quantify the impact of specific item characteristics.
    \item Further refining the LLM-based stimulus generation methods.
\end{itemize}

\section{Conclusion}
\label{sec:conclusion}
This work investigated the impact of stimulus quality on the evaluation of LLM knowledge of complex syntax, focusing on parasitic gaps as studied by Lan et al. (2024). Potential confounds in their stimuli were identified, and it was demonstrated that GPT-2's performance on parasitic gap constructions improves significantly when evaluated on a refined dataset designed to mitigate these issues. 

Preliminary results suggest that conclusions about an LLM's failure to acquire a phenomenon may be premature if based on stimuli with confounds. This underscores the critical importance of stimulus design. The initial results reported here underscore the critical importance of stimulus quality in the evaluation of LLM syntactic abilities and have direct bearing on debates surrounding linguistic nativism and learnability.

\bibliography{custom}

\newpage
\onecolumn
\appendix

\section{Prompt Template for Stimulus Generation}
\label{sec:appendix_prompt}

Prompt for data generation:

\begin{lstlisting}[style=promptstyle, caption={Gemini 2.5 Prompt Template}, label=lst:prompt]

Your task is to generate 10 unique item sets for testing parasitic gap constructions in English. Each item set must consist of exactly four sentences, following a $2 \times 2$ factorial design: +/- Filler and +/- Main Clause Gap (G2). The output should be formatted as a series of comma-separated lines, with each line representing one sentence.

**Objective:**

The primary goal is to create natural-sounding and grammatically clear sentences. The `+Filler, +Gap` sentence in each set must be a canonical parasitic gap construction where the wh-filler "who" is co-indexed with two gaps: G1 (the parasitic gap within a subject NP island) and G2 (the host gap, object of the main embedded verb). This co-indexed reading should be pragmatically plausible. The critical material differentiating the `+G2` (gapped) and `-G2` (filled) conditions for the main clause verb must be a single word.

**Core Sentence Structure for Parasitic Gap (`+Filler, +Gap` condition):**

`[Preamble] who [SubjectNP containing G1] [MatrixVerbPhrase licensing G2] [ADV_Post_G2_Gap].` (Note: The gap for G2 is implied before the ADV_Post_G2_Gap).

**Detailed Constraints for Sentence Components:**

1. **Preamble:** Choose from simple introductory phrases like: "I know", "She heard", "They believe", "The report suggested", "It is clear".
2. **Filler/Complementizer:**
* `+Filler` conditions use: "who"
* `-Filler` conditions use: "that"
3. **Subject NP containing G1 (The Island):**
* This NP must be the subject of the matrix verb phrase. The gap G1 is the object of the preposition.
* Structure: "the `[NounHead]` `[Preposition]`" (The gap G1 is implied after the preposition).
* `[NounHead]`: Use common nouns that naturally take a PP complement with "about" or "of" where the object of the preposition can be a person. Examples: "story", "report", "book", "article", "picture", "critique", "rumor", "discussion", "painting", "description".
* `[Preposition]`: Use **only "about" or "of"**. Select the preposition that forms the most natural phrase with your chosen `[NounHead]`.
4. **Matrix Verb Phrase (licensing G2):**
* Structure: `[LinkingVerb] [TransitiveVerb_G2]` (The gap G2 or object G2_FillerObject follows this).
* `[LinkingVerb]`: Use common linking phrases like: "is likely to", "is going to", "is expected to", "will probably", "might".
* `[TransitiveVerb_G2]`: Use common transitive verbs that naturally take a person as a direct object (for G2). Examples: "upset", "amuse", "delight", "interest", "surprise", "anger", "please", "concern", "bother", "disturb", "fascinate".
5. **Critical Word for +G2 (Gapped) Condition:**
* `[ADV_Post_G2_Gap]`: When G2 is gapped, the sentence should continue immediately after `[TransitiveVerb_G2]` with a single, common adverb from the following list ONLY: "soon", "eventually". This adverb signals the gapped G2.
6. **Lexical Items for Filled Gaps:**
* `[G1_FillerName]` (fills G1 in `-Filler` conditions that also have G1 filled): Use common, simple proper names (e.g., "Mary", "John", "Sarah", "the manager").
* `[G2_FillerObject]` (fills G2 in `-G2` conditions): **Use ONLY a single common proper name from a list such as: "Anna", "Ben", "Chris", "Dana", "Leo", "Sara", "Tom", "Paul", "Nina". Please vary the names used. Avoid using "Kim" for this slot if other simple names from this list or similar common single names are suitable.** The goal is a single-word proper name.
* Ensure `[G1_FillerName]` and `[G2_FillerObject]` are different within the same item set.


**Factorial Design - Sentence Patterns for Each Item Set:**
(Note: Gaps are implied by the structure and absence of overt objects.)

1. **`PFPG` (`+Filler, +G1_gap, +G2_gap`):**
`[Preamble] who the [NounHead] [Preposition] [LinkingVerb] [TransitiveVerb_G2] [ADV_Post_G2_Gap].`
*Example: I know who the story about is likely to amuse soon.*
2. **`MFPG` (`-Filler, +G1_filled, +G2_gap`):**
`[Preamble] that the [NounHead] [Preposition] [G1_FillerName] [LinkingVerb] [TransitiveVerb_G2] [ADV_Post_G2_Gap].`
*Example: *I know that the story about Mary is likely to amuse soon.*
3. **`PFMG` (`+Filler, +G1_gap, -G2_filled`):**
`[Preamble] who the [NounHead] [Preposition] [LinkingVerb] [TransitiveVerb_G2] [G2_FillerObject] [ADV_Post_G2_Gap].`
*Example: *I know who the story about is likely to amuse Anna soon.*
4. **`MFMG` (`-Filler, +G1_filled, -G2_filled`):**
`[Preamble] that the [NounHead] [Preposition] [G1_FillerName] [LinkingVerb] [TransitiveVerb_G2] [G2_FillerObject] [ADV_Post_G2_Gap].`
*Example: I know that the story about Mary is likely to amuse Anna soon.*
**Output Format and Instructions for Generation:**

Please provide 10 unique item sets. For each item set, output four lines, each corresponding to one of the conditions below. Each line must follow this exact comma-separated format:

`sentence_type,item_id,condition,full_sentence`
* **`sentence_type`**: Use the value "subject_pg" for all sentences.
* **`item_id`**: Use a unique integer for each set (e.g., 1 for the first set of four sentences, 2 for the second set, and so on, up to 10).
* **`condition`**: Use the labels "PFPG", "MFPG", "PFMG", "MFMG" respectively for the four sentences in each item set, corresponding to the patterns defined above.
* **`full_sentence`**: The generated sentence string, ending with a period.

**Example of desired output format for ONE item set (item_id 1):**
subject_pg,1,PFPG,I know who the story about is likely to amuse soon.
subject_pg,1,MFPG,I know that the story about Mary is likely to amuse soon.
subject_pg,1,PFMG,I know who the story about is likely to amuse Anna soon.
subject_pg,1,MFMG,I know that the story about Mary is likely to amuse Anna soon.

**Crucial Reminders for Generation:**
* Vary lexical choices for `[Preamble]`, `[NounHead]`, `[Preposition]` (choose 'of' or 'about'), `[G1_FillerName]`, `[LinkingVerb]`, `[TransitiveVerb_G2]`, `[G2_FillerObject]` (from the restricted list of names), and `[ADV_Post_G2_Gap]` (from the restricted list) across the 10 item sets to ensure diversity.
* All `PFPG` sentences must be natural, unambiguously grammatical parasitic gap constructions with a pragmatically plausible co-indexed reading for "who". The subject NP containing G1 must clearly function as a syntactic island.
* All grammatical sentences (PFPG and MFMG) must be clearly grammatical; ungrammatical sentences (MFPG and PFMG) must be clearly ungrammatical due to the specified filler/gap violations.
\end{lstlisting}

\end{document}